\documentclass[12pt,a4paper]{article}
\usepackage{amsmath}
\usepackage{graphicx}
\usepackage{times}
\usepackage{cite}
\usepackage[format=hang]{caption}
\usepackage{tabularray}
\usepackage{xcolor}
\usepackage{multirow}
\begin{document}
\title{A Comparative Analysis of Visual Odometry in Virtual and Real-World Railways Environments}

\author{
    Gianluca D'Amico,
    Mauro Marinoni,
    Giorgio Buttazzo\\
    \textit{Department of Excellence in Robotics \& AI},\\ Scuola Superiore Sant'Anna, Pisa, Italy
}

\maketitle
\section*{Abstract}

This paper aims at illustrating the benefits of employing graphic simulation for early-stage testing of perception tasks in the railway domain, For this purpose, it presents a comparative analysis of the performance of a SLAM algorithm applied both in a virtual synthetic environment and a real-world scenario.
The analysis leverages virtual railway environments created with the latest version of Unreal Engine, facilitating data collection and allowing the examination of challenging scenarios, including low-visibility, dangerous operational modes, and complex environments.
The results highlight the feasibility and potentiality of graphic simulation to advance perception tasks in the railway domain.

\vspace{0.2in}
\noindent{\bf Keywords:} railway automation, simulation framework, railway environments, graphic engines, perception tasks, SLAM algorithms.
\section{Introduction}\label{s:In}

Accurate train localization has paramount importance in ensuring the safety, efficiency, and reliability of railway transportation systems. 
The railway industry faces growing pressure to adopt more advanced and automated operations to address this issue to reduce operating costs and increase line capacity. 
For instance, traditional methods for train localization, primarily reliant on track-mounted transponders and wheel odometers, serve as the backbone of railway operations, providing essential data for navigation and control. 
However, they suffer from significant drawbacks, including high maintenance costs, susceptibility to environmental factors, and the inherent inaccuracies associated with mechanical wear (e.g., sleep and slide phenomena) and environmental conditions. 
These problems can compromise the reliability of train localization and limit the potential for advanced automation and safety enhancements.

The research in the automotive field has introduced many perception-based algorithms capable of guaranteeing a high localization accuracy based on a wide range of sensors, such as RGB cameras, infrared cameras, and Light Detection And Range (LiDARs) sensors, that can also be applied in the railway domain.
Perception tasks, such as visual or LiDAR odometry, Simultaneous Localization and Mapping (SLAM), image segmentation, and object detection, are crucial not only for train localization but also for the automation of various train operations like track discrimination, emergency braking, or signaling.

Also in the railway domain, there is a growing interest in computer vision and deep learning techniques for perceptual tasks, as clearly highlighted by Etxeberria-Garcia et al.~\cite{VOsubway, CVtrainstop}.
However, implementing and testing these algorithms in actual railway environments presents practical challenges, mainly due to the difficulties of equipping trains with the required sensors and the severe restrictions in accessing railway infrastructures.

To overcome such difficulties, in the last years, graphic simulation has been successfully employed in different application domains as a flexible tool for training and testing machine learning algorithms in several scenarios, including corner cases that would be difficult to replicate in a real environment. Examples of such simulators are CARLA~\cite{CARLA} for self-driving cars, AirSim~\cite{AirSim} for unmanned aerial vehicles, and Open AI Gym~\cite{OpenAI Gym} for robot control.
In particular, such simulators enable the automatic generation of large labeled datasets that can speed up the development and testing of novel perception and control algorithms.

In the railway domain, TrainSim~\cite{TrainSim} was developed above Unreal Engine 4~\cite{UE} to generate synthetic scenarios for testing algorithms that exploit images and point clouds to implement tasks for train localization, track discrimination, and pose estimation.
Based on the knowledge gained from TrainSim, we developed a new simulation framework on top of Unreal Engine 5, called RailSim, to generate highly photo-realistic scenarios. Such a new framework was developed from scratch as an independent project and has been optimized for real-time performance and automatic labelling. 

Another simulation framework for generating railway scenarios was developed by de Gordoa et al.~\cite{RailwayCarlaSim} on top of CARLA~\cite{CARLA} and used to generate a public dataset of about $200$ RGB images gathered from 68 different railway scenarios under various lighting and weather conditions. However, such a dataset only comprises semantic segmentation labels and track discrimination masks.
Decker et al.~\cite{GANsyn} exploited Generative Adversarial Networks (GANs) to produce a dataset of synthetic images from authentic ones, primarily focusing on semantic segmentation, while Chen et al.~\cite{RailFOD23} used ChatGPT~\cite{ChatGPT} (Chat Generative Pre-Trained Transformer) and Stable Diffusion models~\cite{DiffusionModel} to detect foreign object on railroad transmission lines.

To the best of our knowledge, none of the existing work focused on train odometry. 
To overcome these limitations, this paper presents a preliminary analysis of RailSim to evaluate its suitability for testing visual odometry algorithms. 
To this end, we compared the results obtained by the ORB-SLAM2 algorithm~\cite{ORB-SLAM2} on both synthetic images generated by RailSim against real-world images taken from the OSDaR23 dataset~\cite{Osdar23} for the monocular camera.
We also carried out experiments on synthetic sequences at different train speeds and image resolutions.
In particular, a comparative analysis of monocular and stereo-based SLAM algorithms has been carried out to show that graphic simulation represents a crucial tool for developing and testing novel train localization algorithms.

The rest of the paper is organized as follows.
Section~\ref{ss:RW} presents an overview of the related literature.
Section~\ref{S:Pr} introduces the proposed approach, briefly illustrating some examples of the generated scenarios and describing the virtual odometry algorithm.
Section~\ref{s:Res} presents the comparative study, reporting the results achieved for the SLAM algorithm under the virtual and real-world data.
Finally, Section~\ref{s:Con} summarizes the conclusions and presents some future work.

\section{Related works}\label{ss:RW}

Since the growing interest in deep neural networks (DNN) and automated systems, public datasets have become essential to test and validate new perception algorithms and compare them over common references.

This approach is already established in the automotive and robotic domains, where various datasets and benchmarks are available for different types of sensors and perception tasks. The KITTI vision benchmark suite~\cite{KITTI} is a public dataset created by equipping a station wagon with several sensors, such as cameras, LiDARs, and odometer. It provides finely created labels for multiple perception tasks, such as visual odometry, object detection, and optical flow.

Similarly, different simulation tools have been developed for creating virtual worlds in different application domains, allowing the user to gather data from several points of view and different environmental conditions, such as lightning or weather changes.
For instance, CARLA~\cite{CARLA} is an open-source simulator based on Unreal Engine 4 (UE4)~\cite{UE} that allows testing self-driving vehicles in a urban environment in which the controlled vehicle can acquire data from different sensors and interact with the virtual world through proper actions.
Similarly, \textit{AirSim}~\cite{AirSim} is a simulation tool based on the UE4 graphic engine tailored for unmanned aerial vehicles (UAV).
Another tool is AutonoVi-Sim~\cite{Autonovi-sim}, which supplies LiDAR frames gathered into a virtual world.

In the railway domain, there is a lack of public datasets of real-world images, due to the difficulties of equipping trains with the required sensors, and the restrictions in accessing the infrastructures.
The most promising public datasets are RailSem19~\cite{RailSem19}, which provides 8500 images for semantic segmentation in railways environment, FRSign~\cite{FRSign}, and GeRARLD~\cite{GERARLD}, which contain several images for traffic lights detection.
To the best of our knowledge, the only datasets that can be exploited for visual odometry are Nordland~\cite{Nordland} and OsDaR23~\cite{Osdar23}.
The former focuses on visual place recognition, providing four different synchronized videos, gathered in the four seasons, along with interpolated GPS raw data.
The latter includes images acquired from several sensors, as RGB and infrared cameras, LiDAR and radar, along with precise odometry data derived from an inertial navigation system based on the integration of GPS with inertial measurements units, thus exploitable for SLAM system evaluation.

Another dataset of synthetic images was presented by de Gordoa et al.~\cite{RailwayCarlaSim}, who exploited an extension of CARLA to generate railway images in different light and weather conditions.
The employment of the CARLA framework allowed the authors not only to generate data and labels for different tasks, such as semantic segmentation masks, 3D bounding boxes, and odometry data, but also to test the effects of light and weather condition changes on the scene.
Unfortunately, the authors did not show the sim-to-real transferring capabilities of the generated data, by training a segmentation network on synthetic images and testing it on real ones under different environmental conditions. Moreover, the public dataset contains only images and does not provide any odometry data to test visual odometry or SLAM algorithms.

Other solutions to gather synthetic data in railway domain include the manipulation of real world images.
For instance, Decker et al.~\cite{GANsyn} generated synthetic data with a GAN to provide a dataset for semantic segmentation task. Instead, Chen et al.~\cite{RailFOD23} employed ChatGPT~\cite{ChatGPT} and Stable Diffusion model~\cite{DiffusionModel} to generate synthetic data of foreign object detection on railroad transmission lines.
However, these datasets do not target visual odometry tasks, thus are not applicable for this study.

As far the simulation frameworks are concerned, TrainSim~\cite{TrainSim} is a simulation framework developed above Unreal Engine 4 to generate virtual railway scenarios for testing perceptual algorithms. 
It can automatically generates labeled datasets of images and point clouds for several visual tasks, such as semantic segmentation, and track discrimination. 
Furthermore, it is possible to extract the poses of the train along the route synchronized with the other sensory data, allowing to test also visual and LiDAR odometry algorithms.
The effectiveness of the simulated LiDAR implemented in TrainSim was validated by a performance study~\cite{LidarTrainSim} aimed at testing not only the validity of the 3D data points, but also the back-scatterted intensity data produced by a Lambertian-Beckmann model.

In conclusion, although the advancements in simulation frameworks and synthetic data generation offer promising avenues for addressing the scarcity of publicly available datasets in the railway domain, further efforts are needed to develop comprehensive datasets that encompass a more complete set of tasks, including visual odometry.

\section{Proposed Approach}\label{S:Pr}

This section describes the proposed approach, presenting a brief overview of the RailSim framework used for the generation of the synthetic images and the ORB-SLAM2 algorithm selected for the evaluation study considered in this paper.

\subsection{Graphic Simulation for Railway Environment}\label{ss:GS}

All the synthetic images used in the evaluation study presented in Section~\ref{s:Res} have been generated by RailSim, a flexible tool for generating virtual railway environments and simulate a set of virtual sensors, as cameras, LiDARs, and inertial measurement units.
The environment is generated according to a number of user-defined configuration parameters concerning the track features, the environment, and the weather conditions. It can also be used to 
automatically create a labeled dataset for several perception tasks, as presented in the paper describing the first prototype of the simulator~\cite{TrainSim}.
Providing all the details about the simulator goes beyond the scope of this paper. Instead, this section provides a brief overview of the main features of the virtual railway scenarios generated to produce the dataset used in the paper. In particular, we present the types of objects included in each scenario and describe the process used for capturing the image sequences.

\begin{figure}[htbp!]
\centering
\includegraphics[width=\columnwidth]{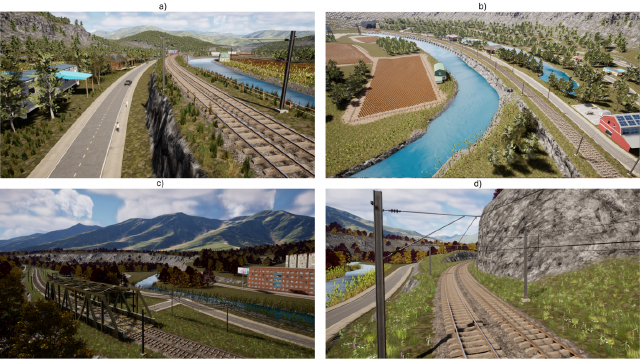}
\vspace{-5mm}
\caption{Samples scenarios created in RailSim: a) urban environment including moving objects; b) farmland/rural areas including agricultural activities; c) bridge; d) detail of the electrified line.}\label{img:virtual_environments}
\end{figure}

Three diverse scenarios have been crafted in RailSim, including the surroundings of the rail-tracks, such as:
\begin{itemize}
    \item Urban areas: scenarios with buildings, streets, and urban infrastructures, as shown in Figure~\ref{img:virtual_environments}.a.
    \item Farmlands: open fields with agricultural activities, providing a rural backdrop, as shown in Figure~\ref{img:virtual_environments}.b.   
    \item Industrial area: open fields with factory activities, presenting warehouses and hangers characteristic of industrial zones.
    \item Grasslands: expansive grass areas typical of natural undeveloped land.
    \item Forests: dense area with trees, representing wooded or forested areas.
\end{itemize}
The type of the surrounding area was guided by the nature of typical railway environments, where rail tracks span mostly rural areas, connecting different towns and running across agricultural, industrial, and natural areas.  

The train routes were taken from OpenStreetMap~\cite{OSM} and integrated into the UE5 virtual world, ensuring terrain conformity.
The railway infrastructures were taken from two UE5 plugins~\cite{TrainTemplate, RailwaySystemTemp} that include the following structure meshes: bridges as shown in Figure~\ref{img:virtual_environments}.c, tunnels, electrified lines, railway crossing, trains, and track-bed including ballast, sleepers, and rails, as shown in Figure~\ref{img:virtual_environments}.d.

To simplify the simulation, the train trajectory was precomputed based on a static velocity profile (gathered from real data acquired in OpenRailwayMap~\cite{OSM} or manually defined), exploiting a first-order system in which the train trajectory follows the static velocity profile without exceeding a fixed maximum allowed acceleration/deceleration. Adopting a more realistic dynamic model for the train is out of the scope of this work and will be part of a future work.

Based on the desired frame rate, the virtual camera was placed along the precomputed trajectory and programmed to acquire images at given timestamps. For example, if the camera frame rate is $50$ $Hz$, the trajectory is sampled at $20$ $ms$ and, at each graphic frame, the train is moved to the new position along the trajectory.
Images are captured by the UE5's UTextureRenderTarget2D\footnote{https://docs.unrealengine.com/5.2/en-US/API/Runtime/Engine/Engine/UTextureRenderTarget2D/}, which allows the user to adjust various camera parameters like lens focal length, aperture, field of view, exposure, lighting settings, and image resolution.

\subsection{Algorithm for Simultaneous Localization and Mapping}\label{ss:VO}

This paper explores the potential of visual simulation to support the development and testing of computer vision algorithms in a railway environment, focusing on visual odometry and monocular/stereo-based SLAM algorithms for their well-known benefits in enhancing train localization methods.
Note that this work avoids using DNN-based solutions due to the inherent domain shift challenges presented by these models when transferring capabilities between synthetic and real-world data.
Instead, the paper relies on a traditional geometric algorithmic approach, which allows for a more straightforward evaluation of the visual simulator's potential without dealing with the complexity of domain adaptation often encountered in DNNs-based methods.

Monocular visual odometry is a technique for estimating the pose (i.e., position and orientation) of a camera through the analysis of consecutive images. 
It identifies and tracks the movement of key features across these images to deduce how far and in which direction the camera has moved. 
SLAM algorithms extend the capabilities of visual odometry by localizing the camera within its environment and constructing a map of that environment in real-time. 
Such a dual process enables an agent to navigate using visual data and understand its surroundings with greater accuracy, continuously updating its location while identifying significant feature landmarks and using them for navigation and orientation.

Among all, the ORB-SLAM2~\cite{ORB-SLAM2} algorithm is recognized as a state-of-the-art SLAM system, which operates with monocular, stereo, and RGB-D cameras, offering real-time functions across a broad spectrum of environments. 
The ORB-SLAM2 methodology unfolds through several critical phases:
\begin{itemize}
    \item Initialization: the algorithm starts by generating an initial map with a few frames, called keyframes, establishing the camera's initial pose and a sparse 3D point cloud, crucial for monocular setups.
    \item Feature Extraction: it detects Oriented FAST and Rotated BRIEF (ORB) features, used as the basis for subsequent operations within the system for their efficiency and robustness.
    \item Tracking: the algorithm estimates the pose of the camera in each new frame by matching ORB features against previous frames or the local map, refining the pose with motion models, and executing relocalization procedures as needed.
    \item Local Mapping: new keyframes contribute to the construction of a local map, incorporating steps like triangulating new points, executing local bundle adjustments, and pruning poorly tracked map points.
    \item Loop Closing: the system identifies revisited areas and performs pose graph optimization for global map consistency upon loop detection to correct drift over time.
    \item Global Optimization: after loop closure, comprehensive bundle adjustment refines the entire map, enhancing the overall accuracy of the system.
    \item Keyframe and Map Point Management: Decisions regarding introducing new keyframes and maintaining map points are made to preserve the most relevant data for real-time application.
\end{itemize}

To summarize, ORB-SLAM2 is a system that integrates feature extraction, motion estimation, both local and global map optimizations, and loop closure detection. This ensemble of components provides precise and efficient SLAM capabilities suitable for a wide range of environments.
The ORB-SLAM2 algorithm is selected for its versatility across various sensor configurations (including monocular, stereo, and RGB-D cameras) and its high reliability and accuracy. 

\section{Experimental results}\label{s:Res}

This section presents the results obtained by applying the ORB-SLAM2~\cite{ORB-SLAM2} algorithm to real-world and synthetic data. In particular, Subsection~\ref{ss:PreConc} introduces some key concepts necessary to understand the testing procedure;
Subsection~\ref{ss:ResVOOsdar} reports the results achieved by the monocular ORB-SLAM2 algorithm applied to real-world images of the OSDaR23 dataset~\cite{Osdar23} and synthetic ones gathered from RailSim; and Subsection~\ref{ss:OtherExp} presents additional experiments on the stereo ORB-SLAM2 performed on RailSim to evaluate the algorithm accuracy in a railway environment.

\subsection{Preliminary concepts}\label{ss:PreConc}

A visual odometry algorithm estimates the 6-DOF (degrees of freedom) poses of a camera directly from an image sequence.
In this work, only the translation part of a pose is considered, due to the almost planar evolution of the train trajectory.
The ORB-SLAM2 algorithm is applied to a sequence $S=\{img_0, img_1, ..., img_n\}$ of $n$ images to estimate a sequence $P=\{p_0, p_1, ..., p_m\}$ of $m$ 3-DOF poses, where $m <= n$, and $p_i = (x_i, y_i, z_i)$ with $i \in [0, m]$.
At the beginning of the image sequence, ORB-SLAM2 starts the map initialization process, which ends when enough high-quality features are found in the images. 
During normal operation, ORB-SLAM2 tracks the features in the next frame and updates the pose; if it is not able to track the features, it resets the keyframe map and restarts the initialization phase.
The performance of the algorithm can be tuned by different parameters, which affect its computational efficiency, robustness, and accuracy.
One of the main parameters is the number of features (i.e., \texttt{nFeatures}) included in the key frame map. 
It defines the maximum number of features to be extracted during initialization and tracking phases.
Increasing this value can be beneficial for applications requiring detailed spatial information but might introduce noise due to false positives among the additional features.

To evaluate the results of ORB-SLAM2, we opted for the APE (Absolute Pose Error) and RPE (Relative Pose Error) metrics. They were chosen to assess the global consistency of the estimated poses relative to ground truth data and quantify the estimation error between consecutive images, respectively.

Finally, the performance of the ORB-SLAM2 algorithm on synthetic images was tested both in monocular and stereo mode in two different virtual scenarios, as described in the following sections.

\subsection{Monocular SLAM compared with OSDaR23}\label{ss:ResVOOsdar}

The OSDaR23~\cite{Osdar23} dataset comprises 45 sequences of annotated multi-sensor data. 
It includes images captured by two distinct groups of cameras, one with a resolution of $4112$ × $2504$ and the other with $2464$ × $1600$, operating at 10 frames per second (fps).
Each group consists of three cameras positioned in a central, diagonal left, and diagonal right alignment. 
Additionally, the dataset contains train poses computed from INS and GPS sensors.
In our analysis, we only used the central RGB mono camera at a lower resolution, because (i) the high-resolution cameras require substantial computational resources that would prevent real-time processing, and (ii) the angle between each camera is too large for depth estimation in the stereo algorithm, also due to the narrow focal length of only $30^\circ$.

Furthermore, we selected only three sequences, containing 100 sensor frames each, because the other sequences include only 10 frames, which are too short for applying visual odometry.
Namely, the ``3\_fire\_site\_3.1'' sequence includes vegetation and fences, the ``5\_station\_bergedorf\_5.1'' sequence contains vegetation and other wagons, and the ``6\_station\_klein\_flottbek\_6.2'' sequence is gathered near a station.

\begin{figure}[htbp!]
\centering
\includegraphics[width=\columnwidth]{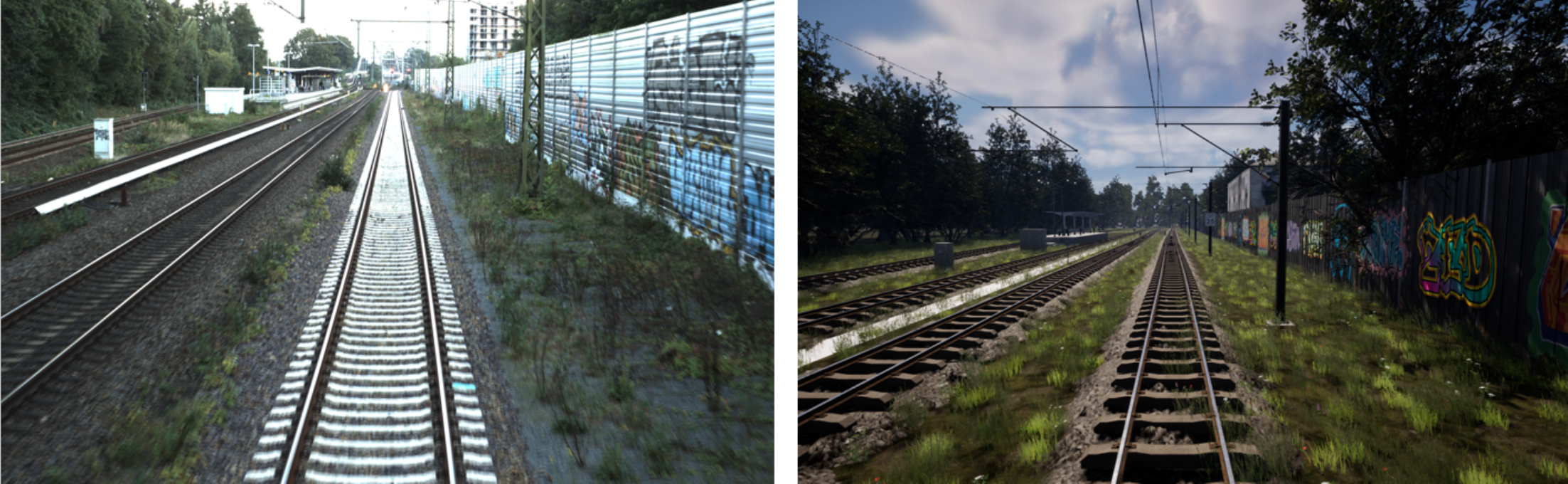}
\vspace{-5mm}
\caption{First image of the OSDaR23 "3\_fire\_site\_3.1" sequence (left) and its virtual replica generated by Railsim (right).}
\label{img:replica}
\end{figure}

To ensure a fair comparison with the real-world data, three synthetic scenarios were generated by RailSim to mimic similar environemnts found in the selected OSDaR23 sequences.
The first scenario, denoted as "Replica", is a virtual replica of the "3\_fire\_site\_3.1" sequence in OSDaR23; they a both depicted in Figure~\ref{img:replica}.
The real train route data were taken from OpenStreetMap (OSM) and replicated by adding similar objects in the scene.
Although the synthetic scenario may differ from the real one, the similarity of the objects present in both scenes allows for a fair comparison of the ORB-SLAM2 algorithm for the purposes of this study.

\begin{figure}[htb]
\centering
\includegraphics[scale=2]{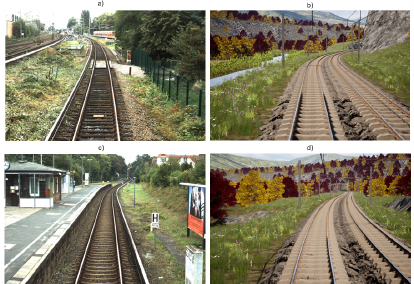}
\vspace{-2mm}
\caption{Images taken from train perspective of: ``5\_station\_bergedorf\_5.1'' OSDaR23 sequence (a); open landscape scenario of ``RailSim\_00'' (b); ``6\_station\_klein\_flottbek\_6.2'' OSDaR23 sequence (c); open landscape scenario of ``RailSim\_01'' (d).}
\label{img:test_sequences}
\end{figure}

The second and third scenario, “RailSim\_00” and “RailSim\_01”, includes a path passing through farmlands and open space landscapes traversed at an average speed of 35 km/h, using a camera with $2464$ × $1600$ pixels, acquiring images at 10 fps, and with the same focal length. They are both illustrated in Figure~\ref{img:test_sequences} in comparison with the real ones.

Noted that, for both synthetic and real-world sequences, the \texttt{nFeatures} parameter was raised from 2000 (default value for the KITTI vision benchmark) to 7000 to compensate for the lower number of features present in typical open-space railway landscapes, and for the differences in focal length.

Table~\ref{t:mono_results} shows the results of ORB-SLAM2 applied to each described sequence in terms of RPE and APE, computed after aligning the resulting trajectories to the ground truth by the Umeyama's algorithm~\cite{umeyama}.

\begin{table}[htbp!]
\centering
\setlength\tabcolsep{3.9pt}
\begin{tabular}{|c|c|rr|rr|}
 \hline
 \multirow{2}{*}{Sequence} & Lenght & \multicolumn{2}{c|}{RPE (m)} & \multicolumn{2}{c|}{APE (m)}  \\
   & (m) & \multicolumn{1}{c}{$\mu\pm\sigma$} & \multicolumn{1}{c|}{Max} & \multicolumn{1}{c}{$\mu\pm\sigma$} & \multicolumn{1}{c|}{Max}  \\ \hline
``3\_fire\_site\_3.1'' & 81.99 & 0.224 $\pm$0.022 & 0.255 & 0.013 $\pm$0.006 & 0.034   \\
``5\_station\_bergedorf\_5.1'' &  101.47 & 1.128 $\pm$0.014 & 1.176 & 0.014 $\pm$0.008 & 0.036   \\
``6\_station\_klein\_flottbek\_6.2'' & 66.39 &0.994 $\pm$0.024 & 1.003 & 0.158 $\pm$0.079 & 0.418   \\ \hline
Replica & 72.50 & 0.412 $\pm$0.023 & 0.135 & 0.121 $\pm$0.056 & 0.257  \\ 
``RailSim\_00'' & 79.20 & 0.019 $\pm$0.012 & 0.041 & 0.561 $\pm$0.263 & 0.992   \\
``RailSim\_01'' & 80.50 & 2.141 $\pm$0.841 & 3.948 & 0.988 $\pm$0.578 & 2.832   \\\hline
\end{tabular}
\caption{\small{Relative pose error (RPE) and absolute pose error (APE) in meters of the ORB-SLAM2 algorithm applied to real-world data of the OSDaR23 dataset and synthetic data generated by RailSim. For each result, the mean $\pm$ standard deviation and the maximum values are reported.}}
\label{t:mono_results}
\end{table}

Note that the errors obtained with the synthetic data are slightly higher than those obtained with OSDaR23, also for the replicated scenario.
We argue that this is due to the reduced number of surrounding objects in the virtual scenario, which reduces the number of matched features and hence decreases the accuracy of the ORB-SLAM2 algorithm.

\subsection{Other experiments with RailSim}\label{ss:OtherExp}

\begin{figure}[htbp!]
\centering
\includegraphics[width=\columnwidth]{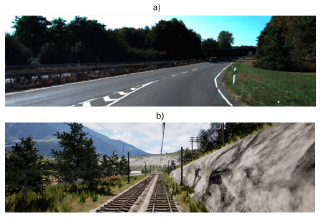}
\vspace{-5mm}
\caption{Images taken from vheicle perspective of: the ``KITTI\_01'' sequence (a); ``RailSim\_U'' synthetic sequence (b).}
\label{img:kitti_ref}
\end{figure}

This section reports another set of experiments aimed at evaluating how the stereo camera affects the performance of the ORB-SLAM2 algorithm.
These tests could not be compared against real-world railway data, since the sequences in the OSDaR23 dataset are not suitable for stereo processing due to the reduced overlapping between the two cameras, as already discussed. For this reason, the results obtained from the synthetic data are compared with a sequence of the KITTI dataset~\cite{KITTI} gathered from a highway as depicted in Figure~\ref{img:kitti_ref}.a, namely ``KITTI\_01'', which better matches a railway scenario.
In this test, the camera resolution and the baseline between cameras were set equal to the ones of the stereo camera system of the KITTI dataset.
The \texttt{nFeatures} parameter for the stereo system required only a minimal increment with respect to the default value (2000) and was set to 2500 (much lower than the value of 7000 used for the monocular algorithm).

The synthetic data were derived from a different scenario, indicated as ``RailSim\_U'', including elements such as buildings, roads, moving objects, and less open landscape with respect to ``RailSim\_00'', and ``RailSim\_01''.
A sample image taken from the train view is shown in Figure~\ref{img:kitti_ref}.b.

The resulting RPE and APE of the ORB-SLAM2 are shown in Table~\ref{t:stereo_results}.
Note that the stereo algorithm does not require the errors to be corrected with Umeyama's algorithm~\cite{umeyama}, since it intrinsically estimates the depth, from which the scale factor can be computed.

\begin{table}[htbp!]
\centering
\setlength\tabcolsep{5pt}
\begin{tabular}{|c|c|rr|rr|}
 \hline
 \multirow{2}{*}{Sequence} & \multirow{2}{*}{Lenght (m)} & \multicolumn{2}{c|}{RPE (m)} & \multicolumn{2}{c|}{APE (m)} \\
     &  & \multicolumn{1}{c}{$\mu\pm\sigma$} & \multicolumn{1}{c|}{Max} & \multicolumn{1}{c}{$\mu\pm\sigma$} & \multicolumn{1}{c|}{Max} \\ \hline
``KITTI\_01'' & 2453.203 & 0.047 $\pm$0.019 & 0.119 & 19.012 $\pm$10.22 & 33.777 \\\hline
``RailSim\_U'' & 809.10 & 0.030 $\pm$0.020 & 0.192 & 1.441 $\pm$0.851 & 3.079 \\\hline
\end{tabular}
\caption{\small{Relative pose error (RPE) and absolute pose error (APE) in meters of the ORB-SLAM2 algorithm applied to the ``KITTI\_01'' sequence and the synthetic data generated by RailSim.}}
\label{t:stereo_results}
\end{table}

The RPE obtained in stereo mode on synthetic data matches closely with the one obtained on the highway scenario of the KITTI dataset.
The considerably higher APE in KITTI is due to the longer sequence (three times the synthetic one), where the pose error accumulates over a larger distance.

Another related issue is that ORB-SLAM2 bases its estimation on both the feature matching and the motion model embedded in the core of the algorithm. 
When the environment contains repetitive features that change with a regular motion, ORB-SLAM2 relies on the motion model to save computational resources.
Hence, a smoother motion of trains along the rails (or vehicles along highways) negatively affects the accuracy of the whole algorithm, with respect to the highly variable motion of cars in the city. 
We argue that taking into account the peculiarities of the train dynamics within the motion model would improve the results of the ORB-SLAM2 algorithm, but such a tuning goes beyond the scope of this paper and will be part of a future work.

\section{Conclusions}\label{s:Con}

This paper presented an evaluation study aimed at showing the validity of synthetic data produced by the RailSim simulation framework for investigating new visual perception algorithms.
The analysis carried out in this work focused on the validation of a state of the art SLAM algorithm, namely ORB-SLAM2, executed on both real-world data, taken from the OSDaR23, and synthetic images generated by RailSim.

The experiments conducted on the monocular system showed similar results between real and synthetic data, providing an evidence of the practical utility of the presented simulation framework.
Also the results obtained on synthetic data with the stereo camera system are comparable with the results obtained in the KITTI sequence on the highway.
Hence, both experiments confirm that the proposed simulation framework represents an essential tool for the development and validation of novel perception algorithms for autonomous vehicles. 
This is especially crucial for railway environments, due to the severe restrictions in accessing railway infrastructures and the difficulties of equipping trains with the required sensors.

As a future work, we plan to modify the motion model of the ORB-SLAM2 algorithm to account for the peculiarities of the train dynamics to further reduce the pose error. 
We will also apply such a type of analysis to other types of algorithms and neural models. 
For this purpose, RailSim is being extended to automatically generate labeled datasets for a range of perception tasks, including object detection, image segmentation, and LiDAR segmentation. Such features will also be essential for training and testing new machine learning models.

\end{document}